\documentclass{article}
\pdfoutput=1

\usepackage{spconf}

\usepackage{amsmath,amsfonts,amssymb}
\usepackage{mathtools}
\usepackage{microtype}
\usepackage{booktabs}
\usepackage{graphicx}
\usepackage{hyperref}
\usepackage{url}
\usepackage{xcolor}

\usepackage{cite}

\usepackage[font=small,skip=1ex]{caption}

\usepackage{multirow}
\usepackage{etoolbox}

\let\citet\cite
\let\citep\cite

\newrobustcmd{\B}{\bfseries}

\title{Textless direct speech-to-speech translation\\ with discrete speech representation}

\name{Xinjian Li\textsuperscript{*\,}\thanks{* Equal contribution. Work done at Google Research.}$^{1,2}$, Ye Jia\textsuperscript{*\,}$^{3}$, Chung-Cheng Chiu$^1$}
\address{
  $^1$Google Research, $^2$Carnegie Mellon University, $^3$Tomato AI \\[4pt]
  \normalsize\texttt{xinjianl\{@google.com,@cs.cmu.edu\}, jiaye@tomato.ai, chungchengc@google.com}}
\begin{document}
\ninept
\maketitle

% \begin{abstract}
% End-to-end speech-to-speech translation (S2ST) without relying on intermediate text representations is a rapidly emerging frontier of research. Recent works have demonstrated that the performance of such direct S2ST systems is approaching that of conventional cascade S2ST when trained on comparable datasets. However, in practice, the performance of direct S2ST is bounded by the availability of paired S2ST training data. In this work, we explore multiple approaches for leveraging much more widely available unsupervised and weakly-supervised speech and text data to improve the performance of direct S2ST based on Translatotron 2. With our most effective approaches, the average translation quality of direct S2ST on 21 language pairs on the CVSS-C corpus is improved by +13.6 BLEU (or +113\% relatively), as compared to the previous state-of-the-art trained without additional data. The improvements on low-resource language are even more significant (+398\% relatively on average). Our comparative studies suggest future research directions for S2ST and speech representation learning.
% \end{abstract}
% \noindent\textbf{Index Terms}: speech-to-speech, speech translation, unsupervised pre-training, multi-task fine-tuning
\begin{abstract}
Research on speech-to-speech translation (S2ST) has progressed rapidly in recent years. Many end-to-end systems have been proposed and show advantages over conventional cascade systems, which are often composed of recognition, translation and synthesis sub-systems. However, most of end-to-end systems still rely on intermediate textual supervision during training, which makes it infeasible to work for languages without written forms. In this work, we propose a novel model, \emph{Textless Translatotron}, which is based on Translatotron~2~\citep{jia2021translatotron}, for training an end-to-end direct S2ST model without any textual supervision. 
Instead of jointly training with an auxiliary task predicting target phonemes as in Translatotron~2, the proposed model uses an auxiliary task predicting discrete speech representations which are obtained from learned or random speech quantizers.
When a speech encoder pre-trained with unsupervised speech data is used for both models, the proposed model obtains translation quality nearly on-par with Translatotron~2 on the multilingual CVSS-C corpus \citep{jia2022cvss} as well as the bilingual Fisher Spanish-English corpus \citep{post2013improved}. On the latter, it outperforms the prior state-of-the-art textless model by $+$18.5 BLEU. % (or $+$58\%).
\end{abstract}

\section{Introduction}

Speech-to-speech translation (S2ST) helps oral communication between people speaking different languages and aims to break such communication barriers. Conventionally, S2ST systems are built with a cascade of automatic speech recognition (ASR), text-to-text machine translation (MT), and text-to-speech synthesis (TTS) sub-systems, all of which rely on intermediate text representations. 
However, the vast majority of the approximately 7,000 languages in the world do not have speech recognition systems or even acknowledged written forms \citep{schultz2006multilingual,nettle2000vanishing, li22aa_interspeech}.
For several widely spoken languages, there are also many regional dialects used for everyday oral communication that differ significantly from formal or standard written forms, such as colloquial Arabic and regional Chinese. S2ST systems relying on intermediate text representations cannot well support such languages.
Additionally, conventional S2ST systems often rely on phoneme representation in TTS and/or ASR. However, for many low resource languages with written forms, there lacks accurate grapheme-to-phoneme conversion tools for building such systems \citep{li2022zero}.

Recently, there has been progress towards developing S2ST systems without relying on intermediate text representations. Such approaches can be put into two categories: 1) End-to-end direct S2ST models, which use a single model to directly translate speech from one language to another \citep{jia2019direct,kano2021transformer,jia2021translatotron,jia2022cvss,jia2022leveraging,dong2022leveraging,shankarappa2022faster}; 2) Cascaded S2ST based on discrete speech representation instead of text \citep{tjandra2019speech,zhang2020uwspeech,lee2021direct,ma2021direct,lee2021textless,huang2022transpeech,popuri2022enhanced}. Although these approaches do not rely on textual representation at inference time, many of them still need textual supervision at training time for obtaining the best performance. A few works have demonstrated the feasibility of training S2ST system without textual supervision, they significantly underperform similar approaches when textual supervision is used \citep{tjandra2019speech,zhang2020uwspeech,lee2021direct}.

In this paper, we propose a novel approach for training end-to-end direct S2ST model without textual supervision. 
%The proposed method combines the best from end-to-end direct S2ST models and discrete speech representations-based cascade models. 
The proposed model, \emph{Textless Translatotron}, is based on Translatotron~2~\citep{jia2021translatotron}, which is an end-to-end direct S2ST model composed of a speech encoder, a linguistic decoder, and an acoustic synthesizer. Instead of predicting the target phonemes as an auxiliary task from the linguistic decoder in Translatotron 2, the linguistic decoder in the proposed model predicts discrete representations of the target speech, which are obtained from a speech quantizer based on VQ-VAE \citep{oord2017neural}. Such discrete speech representations are expected to capture phoneme-like information but without explicitly depending on it. Unlike previous works using discrete representations~\cite{lee2021textless, zhang2020uwspeech}, which require multiple separately trained models (e.g: translation model and vocoder) to be cascaded, our proposed model can be trained end-to-end.
% we further initialize the speech encoder and acoustic synthesizer with the pre-trained models respectively, which significantly improves the learning efficiency.

Experiments on two datasets, including a bilingual dataset and a multilingual dataset, show that our proposed model obtained a very close translation quality compared with the original Translatotron~2, despite of not using textual supervision. Such results significantly outperform the prior state-of-the-art textless model ~\citep{lee2021textless} on the Fisher Spanish-English dataset by $+$18.5 BLEU (or, $+$58\% relatively). %\Ye{should be 18.5 BLEU, right?} % \Ye{mention random quantizer}

\section{Related works}

% \paragraph{Direct S2ST}

Conventional cascade S2ST systems relying on
%text as the intermediate representation
intermediate text representation
are unable to support languages without written forms, or when textual labels are missing from datasets for written languages. The recently emerging research on S2ST without going through intermediate text representation started to explore such scenarios.

The first proposed direct S2ST model, Translatotron \citep{jia2019direct}, used a sequence-to-sequence model with attention to directly translate speech spectrogram in one language to speech spectrogram in a different language. Although it did not rely on textual intermediate representation at inference time, it required auxiliary objectives based on text at training time to obtain reasonable quality. % for the difficulty on learning speech understanding and alignment on long sequences at the same time.

Tjandra et al. \citet{tjandra2019speech} first demonstrated non-trivial results on training S2ST models without any textual supervision at training time, by using a learned discrete speech representation instead of text in a cascade system. It first trained a speech quantizer based on VQ-VAE with a speech spectrogram reconstruction task in a self-supervised manner. The trained VQ-VAE encoder was used for converting the S2ST target speech into a discrete representation. It then trained a second model for translating the S2ST source speech into the discrete representation corresponding to the target speech. The VQ-VAE decoder was used for converting the predicted discrete speech representation into speech spectrograms. The resulting system showed reasonable translation quality, but significantly underperformed baseline systems using text as the intermediate representation.

Improvements on top of \citet{tjandra2019speech} have been primarily focused on learning better discrete speech representations, such as utilizing more training data including labeled data \citep{zhang2020uwspeech}, different learning objective \citep{lee2021direct}, and adopting data augmentation for removing non-linguistic variance such as speaker identity from the learned representation \citep{lee2021textless,huang2022transpeech}.

% \citep{zhang2020uwspeech,lee2021direct,lee2021textless,huang2022transpeech} further improved such an approach, primarily by improving the learned discrete speech representation.

Besides discrete speech representation, continuous speech representations learned on unsupervised data \citep{oord2018representation,baevski2020wav2vec,hsu2021hubert,chung2021w2v,chiu2022self} have also been shown effective in S2ST \citep{jia2022leveraging,popuri2022enhanced}. Such approaches can be naturally adopted for textless S2ST as well.

Our work combines end-to-end direct S2ST with discrete speech representation to benefit from both approaches, and utilizes self-supervised learned continuous speech representation for obtaining best performance. 

% Ye: I don't think S2U+U2S is more relevant than ealier works such as \citet{tjandra2019speech}
% The most relevant work to ours is S2U+U2S~\citep{lee2021direct}, which consists of two sub-systems: the first S2U sub-system takes source speech as input and produces the target HuBERT discrete ids, then the other U2S sub-system utilizes a  Hifi-GAN as a vocoder to generates target speech from those ids~\citep{kong2020hifi}. On the other hand, our model uses VQ-VAE as discrete representations and is a fully end-to-end system without producing any intermediate discrete ids during inference. Additionally, we demonstrate using a pre-trained encoder improve the performance significantly.

%A few works have investigated the model without text supervision \citep{zhang2020uwspeech, lee2021direct}. Most of them depend on the using discrete speech representation instead of text. UWSpeech pre-train its discrete representations utilizing more training data including labeled data from other written languages~\citep{zhang2020uwspeech}, 

%Both resulting systems showed reasonable translation quality, but significantly underperformed baseline systems using text as the intermediate representation.

%Our exploration on random speech quantization is inspired by \citet{chiu2022self}.

\section{Textless Translatotron}
\label{sec:method}

The proposed model, Textless Translatotron, follows the architecture of Translatotron 2 \citep{jia2021translatotron}, which is an end-to-end direct S2ST model. The main components of Translatotron 2 are a speech encoder, a linguistic decoder and an acoustic synthesizer.
% In addition to the translation model, a speech quantizer based on VQ-VAE is pre-trained to extract discrete ids and initialize the synthesizer.
In addition to them, we introduce a speech quantizer based on VQ-VAE, to extract discrete representation from the target speech, which is used for guiding the training of the linguistic decoder.

% A speech quantizer is added in addition to them in this work.

\begin{figure}[t]
    \centering
    \includegraphics[width=0.94\linewidth]{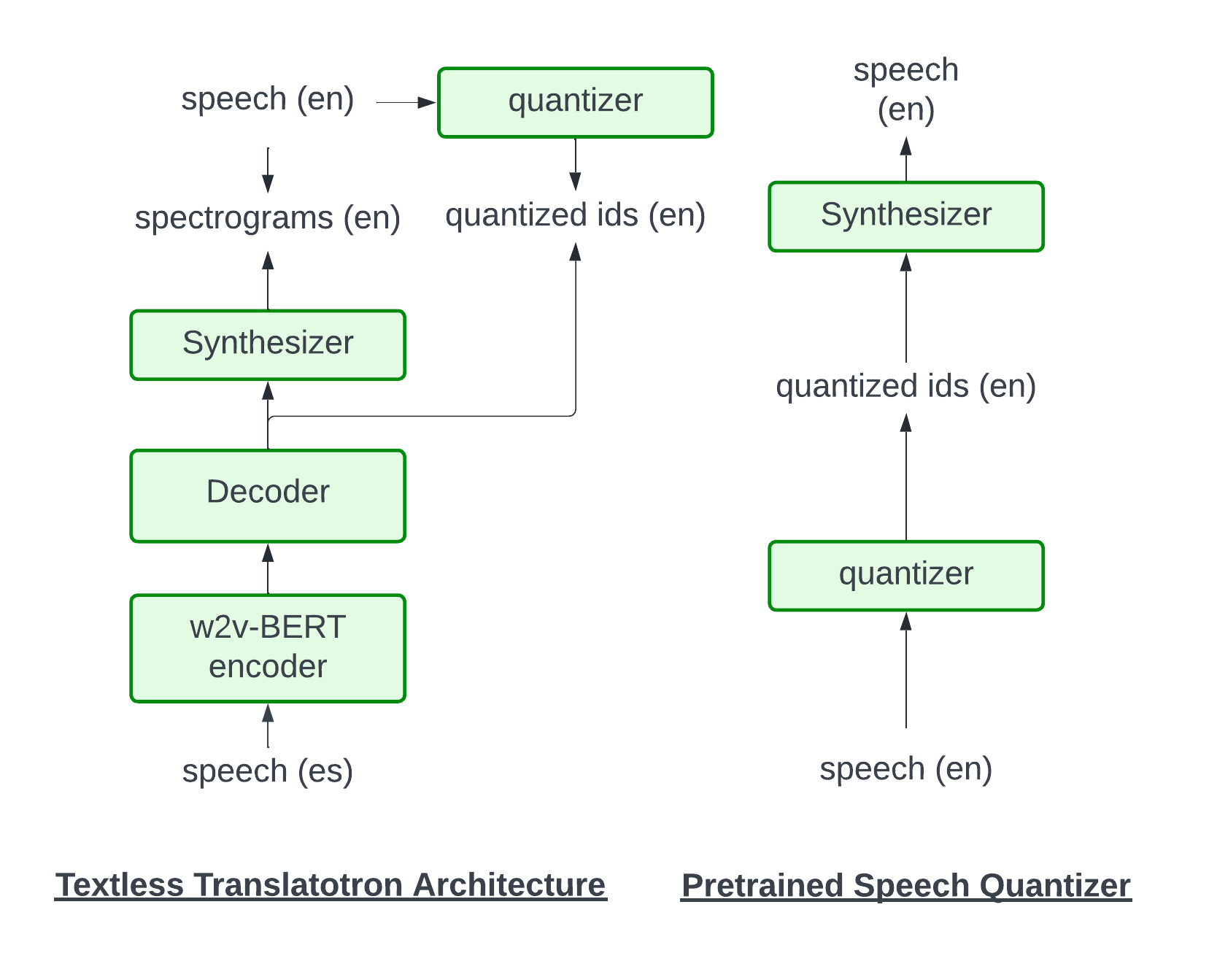}
    \caption{Architecture of our proposed model. The left-side illustrates the speech-to-speech translation architecture: it uses a pre-trained speech encoder and linguistic decoder to output quantized speech ids, which are obtained from a separately pre-trained VQ-VAE model shown on the right side.}
    \label{fig:s2st}
\end{figure}

We improve the Translatotron 2 model by addressing two data scarcity issues: 1) Both the encoder and the decoder were learned from scratch, which is ineffective when training data is scarce. 2) The written form of the target language might be unavailable, therefore supervision using text information becomes impossible. To alleviate those scarcity issues, we improve each of the three components and further add a discrete speech quantizer.

\subsection{Speech encoder}

Instead of training a speech encoder from scratch, we initialize our encoder from a pre-trained multilingual w2v-BERT model \citep{chung2021w2v} as described in \citet{bapna2022mslam}. w2v-BERT is a self-supervised model which combines both contrastive learning and masked language model (MLM) \citep{chung2021w2v}. The w2v-BERT model is pre-trained on a large collection of multilingual speech datasets. It is used to initialize our encoder to extract continuous speech representations and gets fine-tuned during training process.

%It consists of three modules: the feature encoder transforms speech signal into local latent features, the contrastive module is to discretize latent feature into a finite set of speech tokens by solving the contrastive learning task, the discrete features are further processed by the masked prediction module, which is optimized with the MLM objective. %\XJ{reduce the description about w2v-BERT since it is not main focus}

%The w2v-BERT model is pre-trained on a large collection of multilingual speech datasets. It is used to initialize our encoder and gets fine-tuned during training process.

\subsection{Speech quantizer}
\label{sec:quantizer}

To explore discrete speech representations, we consider using VQ-VAE-based speech quantizers as shown on the right-side in Figure~\ref{fig:s2st}. The motivation of choosing VQ-VAE as the quantizer is  that discrete representation learned from VQ-VAE are directly optimized for speech spectrogram reconstruction, which matches how such discrete presentation is used for generating translation speech in S2ST models. In contrast, discrete representation obtained from other models (e.g: HuBERT~\citep{hsu2021hubert}) may not be optimized for spectrogram reconstruction. Additionally, because of such matching, the decoder of the VQ-VAE quantizer can be directly used as the synthesizer in the S2ST models.

The speech quantizers are pre-trained only using the S2ST speech data of the target language (e.g: English only). Let $x$ denote the speech input, the encoder projects it into a latent space $\mathrm{enc}(x)$. Each encoder has a stride hyperparameter which controls the number of frames encoded into each latent vector. We vary the stride from 2 to 16 in our experiment. The model then maps the latent vector to discrete ids through finding a nearest vector in a codebook $C=\{ c_1, c_2, ..., c_n \}$ where $n$ is the codebook size:
\begin{equation}
    c_y = \mathrm{argmin}_i \left\| L^2(c_i) - L^2(\mathrm{enc}(x)) \right\|_2
\end{equation}

Both the codebook and the projected vector are $L^2$ normalized. The decoder takes the discretized representations $c_y$ and attempts to reconstruct the speech inputs $\mathrm{dec}(c_y)$. The reconstruction loss is the absolute difference between the speech input and reconstructed input. Combined with the quantization loss, the total training objective is defined as follows:

\begin{equation}
L = \left\| x - \mathrm{dec}(c_y) \right\|_1 + \alpha \left\|\mathrm{sg}[\mathrm{enc}(x)] - c_y \right\|_2 + \beta \left\| \mathrm{enc}(x) - \mathrm{sg}[c_y] \right\|_2,
\end{equation}
where $\mathrm{sg}$ denotes the stop-gradient operator, $\alpha$ and $\beta$ are hyperparameters which we fix to 1.0 and 0.25, respectively. In our experiments, we adopt a stack of non-causal transformer layers as the decoder following \citep{dosovitskiy2020image}. However, we consider two different groups of quantizers as described in the following subsections.

\subsubsection{Random quantizer}

The random quantization is inspired by the BEST-RQ work~\citep{chiu2022self}. Suppose $[x_1, x_2, ..., x_T]$ denotes the speech input frames where $x_i \in \mathbb{R}^d$ represents the input feature of frame $i$. A stacking process is first applied by combining $s$ frames together without overlapping where $s$ is the stride hyperparameter. The speech input becomes $[x'_1, x'_2, ..., x'_{\lfloor T/s \rfloor} ]$ where $x'_i \in {\mathbb{R}}^{ds}$. The stacked spectrogram is then mapped into a latent space with a projection matrix $A$, i.e. $\mathrm{enc}(x'; A) = Ax'$. Both the projection matrix and the codebook are randomly initialized and fixed during training, only the decoder is optimized. The speech spectrogram is channel-normalized into Gaussian distribution based on global statistics. The projection matrix $A$ is Xavier~\citep{glorot2010understanding} initialized, and the codebook $C$ is Gaussian initialized. Such initialization ensures uniform distribution of the projected code IDs.

%Suppose $[x_1, x_2, ..., x_T]$ denotes the speech spectrogram of an utterance where $x_i \in \mathbb{R}^d$ represents the input feature of timestamp $i$. A stacking process is first applied to the spectrogram: every $s$ frames are stacked together without overlapping where $s$ represents the number of stride. The input spectrogram becomes $[x'_1, x'_2, ..., x_{\lfloor T/s \rfloor} ]$ where $x'_i \in {\mathbb{R}}^{d \times s}$ is the stacked input. 

\subsubsection{Learned quantizer}

To compare the random projection encoder with learned encoders, we explore two more quantizers: a linear quantizer and a Transformer quantizer. The linear quantizer shares the same encoder architecture with the random quantizer, except that the projection matrix $A$ and the codebook $C$ are learned. The transformer encoder is similar to the transformer decoder, which has a stack of non-causal transformer layers. It also stacks $s$ frames before the transformer layers. These learned quantizers are pre-trained on the target speech in the S2ST datasets, and frozen during the training of the translation models.
%\Ye{explain stride and what exactly happends before the Transformer}

\subsection{Linguistic decoder}

%\Ye{TODO: revise this}

Instead of using textual supervision, we use the discrete speech representations from the speech quantizer to guide the training of the linguistic decoder. The linguistic decoder autoregressively predicts the discrete code IDs generated from the speech quantizer.

\subsection{Acoustic synthesizer}

The synthesizer of Textless Translatotron also gets simplified compared to Translatotron 2.
In Translatotron 2, a duration predictor and a Gaussian-weighted upsampler are used to augment the time rate of linguistic representation from the linguistic decoder to match the same of the target speech spectrogram. In Textless Translatotron, because the discrete speech representations have a fixed time rate, a duration predictor is no longer needed, and a simple transposed convolution is used to match the length of the linguistic representation sequence and the spectrogram frame sequence.
Additionally, the autoregressive LSTM stack in the synthesizer of Translatotron 2 is replaced by a non-autoregressive Transformer stack, optionally initialized from the decoder of the learned VQ-VAE quantizers (Sec.~\ref{sec:quantizer}). 

%\Ye{is it initialized or not, when the random quantizer is used?}

\section{Experiments}

To evaluate the effectiveness of the proposed model and the variations described in Sec.~\ref{sec:method}, we conducted comparative experiments on the multilingual CVSS-C corpus~\citep{jia2022cvss} and the bilingual Fisher Spanish-English corpus~\citep{post2013improved}. The CVSS-C corpus contains sentence-level paired S2ST data in 21 X$\to$English language pairs. The source speech in the corpus is 1,153 hours of human read speech collected via crowdsourcing; the target speech in the corpus is 719 hours of high-quality TTS synthetic speech in a single speaker's voice, with speech naturalness on-par with human recordings. The target speech is shorter than the source speech because of better fluency in the TTS synthetic speech. The Fisher Spanish-English corpus contains 127 hours of Spanish telephone conversations and 96 hours of synthetic English translation speech in a single speaker's voice.

All the models are implemented in the Lingvo framework \citep{shen2019lingvo}. Unless specified otherwise, all the S2ST models followed the hyper-parameters from \citet{jia2022leveraging}. The speech quantizer used a 64 dimensional latent space with a codebook size of 512. A 256$\times$12 non-causal Transformer stack is used as the VQ-VAE decoder and the S2ST acoustic synthesizer. The linear-based speech quantizer used a single-layer linear projection, and the Transformer-based speech quantizer used the same 256$\times$12 Transformer stack analogous to its decoder. The pre-trained speech encoder is the same 0.6B-parameter w2v-BERT model from \citep{jia2022leveraging,bapna2022mslam}, which was pre-trained on 492k hours of unlabeled speech in 51 languages.

Following \citet{jia2019direct,jia2022cvss}, the translation quality of S2ST is evaluated by BLEU on ASR transcription from the translation speech (in lowercase, excluding punctuation marks). We used an ASR model from \citet{park2020improved} for evaluation, which is the same as used in \citet{jia2022cvss,jia2022leveraging,jia2021translatotron}, therefore the results are comparable to these works. The results on CVSS-C are grouped into high/mid/low-resource language pair groups based on the amount of data available in the CVSS-C corpus, following \citet{babu2021xlsr,bapna2022mslam}. 

Two groups of baseline models are used for comparison:
text-supervised models and textless models. For the text-supervised models, we refer to the Translatotron~2 models described in \citet{jia2021translatotron,jia2022cvss} and its improved version with the pre-trained encoder~\citep{jia2022leveraging}, which are the state-of-the-art models. For the textless models, we compare our results with UWSpeech~\citep{zhang2020uwspeech} and the prior state-of-the-art in \citep{lee2021direct}.

\subsection{Fisher Spanish-English}

\begin{table}[t]
  \centering
    \caption{Translation quality on Fisher Spanish-English, measured by 4-reference BLEU on the test set.}
  \begin{tabular}{lr}
    \toprule
     & BLEU  \\
    \midrule
    \emph{Textless models} \\
    \hspace{3mm}UWSpeech VQ-VAE~\citep{zhang2020uwspeech} & 3.4 \\
    \hspace{3mm}UWSpeech XL-VAE~\citep{zhang2020uwspeech} & 9.4 \\
    \hspace{3mm}S2U + U2S~\citep{lee2021direct} & 31.8 \\
    \hspace{3mm}Textless Translatotron (this work) & \B{50.3} \\ 
    \midrule
    \emph{Text-supervised models}  \\
    \hspace{3mm}Translatotron \citep{jia2019direct} & 26.9 \\
    \hspace{3mm}S2U + U2S~\citep{lee2021direct} & 39.9 \\
    \hspace{3mm}Translatotron 2 \citep{jia2021translatotron} & 42.4 \\
    \hspace{3mm}Translatotron 2 w/ pre-trained encoder & \B{52.2} \\
    \midrule
    Reference & 88.6 \\
    \bottomrule
  \end{tabular}
  \label{fisher_table}
\end{table}

The experimental results on the Fisher Spanish-English corpus is shown in Table~\ref{fisher_table}, compared to multiple baseline models. Textless Translatotron obtained translation quality approaching the state-of-the-art text-supervised S2ST model Translatotron 2, with a difference of merely 1.9 BLEU. It outperformed the prior state-of-the-art textless S2ST model \citep{lee2021direct} by $+$18.5 BLEU (or $+$58\% relatively).

\subsection{CVSS-C}

\begin{table}[t]
  \centering
    \caption{Translation quality on CVSS-C, measured by average BLEU on 21 X$\to$En language pairs, grouped into high/mid/low-resource language pairs.}
    \setlength{\tabcolsep}{0.5em}
  \begin{tabular}{lrrrr}
    \toprule
     & All & High & Mid & Low \\
    \midrule
    \emph{Textless models} \\
    \quad Textless Translatotron (this work) & \B{17.7} & 33.5 & 22.8 & 10.2 \\ 
    \midrule
    \emph{Text-supervised models}  \\
    \quad Translatotron 2 \citep{jia2022leveraging} & 10.1 & 26.9 & 14.2 & 2.8\\
    \quad\quad w/ pre-trained encoder \citep{jia2022leveraging} & \B{17.9} & 32.5 & 22.9 & 10.9 \\
    \midrule
    Reference & 91.1 & 88.4 & 89.2 & 92.8 \\
    \bottomrule
  \end{tabular}
  \label{cvss_table}
\end{table}

The Fisher Spanish-English corpus contains translation between two close languages and is unable to assess more complicate translation scenarios such as involving heavy re-orderings between languages. 
To further evaluate the performance of the proposed model, we conducted experiments on the multilingual CVSS-C corpus, as shown in Table~\ref{cvss_table}. Textless Translatotron obtained translation quality nearly on-par with the Translatotron 2 model with a pre-trained encoder, with merely 0.2 BLEU difference. 
% Ye: this is likely just difference balancing among languages.
% Compared to By dividing into all languges into high/mid/low resources, we find the textless model has even better performance over the high resources languages (fr, de, ca, es).

\subsection{Ablation studies}

To understand the impact of the designing and hyperparameter choices in the proposed model, we conducted multiple ablation studies on the CVSS-C corpus, as shown in Table~\ref{ablation}, \ref{stride} and \ref{codebook}.

% \subsubsection{Speech quantizer choices}

\subsubsection{Linguistic training targets}

Table~\ref{ablation} shows the impact of the different training target choices for the linguistic decoder.
When the Textless Translatotron is trained without using a pre-trained speech encoder, it underperformed Translatotron 2 signicantly, which used phoneme-based textual supervision. There was no significant performance difference among the quantizer choices, including random quantizer and learned quantizer.
However, when a powerful large pre-trained speech encoder was used, using a learned quantizer, especially one with a larger capacity, showed significantly advantages over a random quantizer or a tiny learned quantizer. With a relatively small Transformer quantizer (256$\times$12), the performance of Textless Translatotron is nearly on-par with Translatotron 2. It is important to note that no extra data other than CVSS-C was used for training the speech quantizer of Textless Translatotron.
These results suggest that one major difficulty on training end-to-end direct S2ST models lies in speech understanding, which can be overcome by either introducing extra explicit supervision as in Translatotron~1~\&~2, or by leveraging self-supervised speech representation learning, as in Textless Translatotron.

\begin{table}[t]
  \centering
    \caption{Ablation studies on the training targets of the linguistic decoder, with or without a pre-trained speech encoder. Reporting average BLEU on all the 21 X$\to$En language pairs in CVSS-C. (Random: random speech quantizer; Linear/Transformer: learned linear or Transformer speech quantizer.)}
  \setlength{\tabcolsep}{0.4em}
  \begin{tabular}{lrrrr}
    \toprule
    \multirow{2.5}{*}{Encoder} & \multicolumn{3}{c}{Textless Translatotron} & Translatotron 2 \\
    \cmidrule(r){2-4} \cmidrule(l){5-5}
   % & Random quantizer & \multicolumn{2}{c}{Learned quantizer} \\
    % \cmidrule(r){2-2} \cmidrule(l){3-4}
     & Random & Linear & Transformer & Phoneme \\
    \midrule
    From scratch & 6.4 & 7.2 & 5.5 & 10.1 \\
    Pre-trained & 13.9 & 16.3 & \B{17.7} & \B{17.9} \\ 
    \bottomrule
  \end{tabular}
  \label{ablation}
\end{table}

\subsubsection{Speech quantizer stride and codebook size}

% \begin{table}[tbh]
%   \centering
%     \caption{compare quantizer}
%   \begin{tabular}{lrrrr}
%     \toprule
%      & All & High & Mid & Low \\
%     \midrule
%     BEST-RQ & 13.9 & 28.3 & 18.4 & 7.2 \\
%     MLP Quantizer & 16.3 & 32.2 & 21.1 & 9.0 \\ 
%     ViT Quantizer & 17.5 & 33.4 & 22.5 & 10.1\\
%     \bottomrule
%   \end{tabular}
%   \label{quantizer}
% \end{table}

% \begin{table}[t]
%   \centering
%     \caption{Translation quality on CVSS-C measured by BLEU, Comparison on the quantizer strides.}
%   \begin{tabular}{rrrrr}
%     \toprule
%     Stride & All & High & Mid & Low \\
%     \midrule
%     2 & 17.3 & 32.6& 22.4 & 10.0 \\
%     4 & 17.5 & 33.4 & 22.5 & 10.1 \\ 
%     8 & 17.1 & 31.8 & 22.0 & 10.1 \\
%     16 & 13.8 & 27.3 & 18.3 & 7.5 \\
%     \bottomrule
%   \end{tabular}
%   \label{stride}
% \end{table}

\begin{table}[t]
  \centering
    \caption{Ablation study on speech quantizer stride. Reporting average BLEU on all the 21 X$\to$En language pairs in CVSS-C.}
  \begin{tabular}{lcccc}
    \toprule
    Stride & 2 & 4 & 8 & 16 \\
    \midrule
    BLEU & 17.3 & 17.5 & 17.1 & 13.8 \\
    \bottomrule
  \end{tabular}
  \label{stride}
\end{table}

% Similarly, varying the codebook size between 128 to 1024 does not huge difference. Increasing the codebook size to 8192 is too fine-grained, which tends to overfit to the dataset and decrease the performance.

% \begin{table}[t]
%   \centering
%     \caption{Translation quality on CVSS-C measured by BLEU. Comparison on the quantizer codebook sizes.}
%   \begin{tabular}{rrrrr}
%     \toprule
%     Codebook & All & High & Mid & Low \\
%     \midrule
%     128 & 17.5 & 33.4 & 22.5 & 10.1 \\
%     512 & 17.7 & 33.5 & 22.8 & 10.2 \\ 
%     1024 & 17.4 & 33.1 & 22.8 & 10.0 \\
%     8192 & 16.5 & 32.1 & 22.0 & 9.1\\
%     \bottomrule
%   \end{tabular}
%   \label{codebook}
% \end{table}

\begin{table}[t]
  \centering
  \caption{Ablation study on speech quantizer codebook size. Reporting average BLEU on all the 21 X$\to$En language pairs in CVSS-C.}
  \begin{tabular}{lcccc}
    \toprule
    Codebook & 128 & 512 & 1024 & 8192 \\
    \midrule
    BLEU & 17.5 & 17.7 & 17.4 & 16.5 \\
    \bottomrule
  \end{tabular}
  \label{codebook}
\end{table}

% \begin{table}[t]
%   \centering
%     \caption{Examples for quantizer stride comparison.}
%   \begin{tabular}{ll}
%     \toprule
%     REF & it's native to northern turkey and northern syria. \\
%     Stride 4 & it is originative to northern turkey and northern syria. \\
%     Stride 16 & it is made from more than turkey and more than syria. \\
%     \bottomrule
%   \end{tabular}
%   \label{quantizer_comparison}
% \end{table}

The stride and the codebook size are two critical hyperparameters of the speech quantizer.
Using a smaller codebook brings stronger supervision to the linguistic decoder, but suffers from larger information loss. Similarly, using a larger stride makes the discrete representation shorter and the training and inference faster, but also suffers from larger information loss.
These two hyperparameters need to be choosed carefully in balancing quality, convergency, and efficiency.

We compared different choices on these two hyperparameters in Table~\ref{stride} and \ref{codebook}. The impact of them are as expected above. We found that using codebook size 512 and stride 4 performed best in the experiments.

% Finally, we fix the Transformer-based quantizer and investigate the impact of varying stride and codebook configurations in Table~\ref{stride} and Table~\ref{codebook}. In Table~\ref{stride}, it shows that the stride size 4 performs best but using the stride size of 2, 4, 8 has no big difference. Further increasing the stride 16 observes huge performance reduction. The reason is that 16 strides typically contain more than 1 phoneme, which makes it difficult to encode much more information into a single token. Similarly, decoding speech signals from such tokens also becomes ambiguous as they have to decide which phoneme to emit. By investigating the pre-training losses, we find the quantization loss among all stride models are similar, however, the reconstruction loss $L_{\text{reconstruct}}$ of the 16 stride model is much higher than the others. This indicates that the large stride has difficulty to decode/reconstruct speech signals. Table~\ref{quantizer_comparison} shows translated examples from a small stride model (4) and a large stride model (16). The small model correctly translate the word "northern" but the large model fails to pronounce it clearly and get recognized as "more than".

\subsection{Sample analysis}

\begin{table}[t]
  \centering
    \caption{Sample translations from Textless Translatotron on CVSS-C. (REF: transcripts of the reference translation speech; HYP: transcripts of the predicted hypothesis translation speech.)}
    \setlength{\tabcolsep}{0.5em}
  \begin{tabular}{ll}
    \toprule
    \multicolumn{2}{l}{\hspace{-0.5ex}\emph{ja$\to$en (low-resource)}}  \\
    REF & everyone knows mount fuji. \\
    HYP & the fuji is long to mina all words. \\
    \midrule
    \multicolumn{2}{l}{\hspace{-0.5ex}\emph{pt$\to$en (mid-resource)}}  \\
    REF & a man and a white dog are looking at a postcard exhibit \\
    HYP & a man in a white dog is looking at a postcards exhibit \\
    \midrule
    \multicolumn{2}{l}{\hspace{-0.5ex}\emph{fr$\to$en (high-resource)}}  \\
    REF & after a year spent in the kibbutz his family arrived in paris \\
    HYP & after a year in the cabot his family arrived in paris \\
    \bottomrule
  \end{tabular}
  \label{cvss_sample}
\end{table}

To understand the failure patterns, we manually analyzed samples of failure cases in the BLEU evaluation. Table~\ref{cvss_sample} cherrypicks a few examples that were considered as failures in such evaluation. One common pattern is that the model does not translate part of the source speech but copies the pronunciation into the prediction without translation. Such direct copying can be desired for words that do not need to be translated, such as names and proper nouns (e.g. ``kibbutz'' in the fr$\to$en example; transcribing to ``cabot'' is an ASR error in the evaluation), as also pointed out in \citep{jia2019direct,jia2022cvss}. However, on low-resource languages, such copying are often real failure cases (e.g. in the ja$\to$en example, ``mina'' means ``everyone'' in Japanese). Such failure cases can likely be improved by having more training data.

% in each of the three groups: In the Low-resource language example, the model can occasionally identify a few words correctly, but has difficulty in understanding the syntactic structure of the sentence. Even with the pre-trained componenets, we find the model still suffers from the scarcity of training data and might not learn enough correlation among words. For instance, we observe it tends to copy the original pronunciation without translation, in our example, it copies and pronounces "mina" (which means everyone in Japanese) without translation. In the Mid-resource group, it have much better understanding of the sentence structure, however, it still have some grammatical issues (in our example it fails to identify the subject). In the High-resource language example, it is more likely to capture the meaning of sentence, though it might fail to translate a few proper noun correctly (Kibbutz in our example)

\section{Conclusions}
We proposed Textless Translatotron, a novel end-to-end S2ST model that can be trained without any textual labels, therefore supports languages without written forms. The proposed model is based on Translatotron 2, but uses discrete speech representation obtained from a VQ-VAE quantizer instead of phonemes to guide the training of the linguistic decoder. When a large pre-trained speech encoder is used in both the proposed model and the baselines, Textless Translatotron demonstraded performance nearly on-par with the state-of-the-art direct S2ST model with textual supervision on the bilingual Fisher Spanish-English corpus and the multilingual CVSS-C corpus, and outperformed the prior state-of-the-art textless S2ST model on Fisher Spanish-English by $+$18.5 BLEU (or $+$58\% relatively).

\section{Acknowledgments}

The authors thank Ankur Bapna, James Qin and Yonghui Wu for helpful discussion and feedback.

% \bibliographystyle{IEEEtran}
% \bibliography{mybib}

\vfill\pagebreak

\bibliographystyle{IEEEbib}
\bibliography{references}

% % \newpage
% % \input{9-appendix}

\end{document}